\documentclass[10pt,twocolumn,letterpaper]{article}

\usepackage[pagenumbers]{iccv} 
\usepackage{multirow}
\usepackage{subcaption}
\definecolor{iccvblue}{rgb}{0.21,0.49,0.74}
\usepackage[pagebackref,breaklinks,colorlinks,allcolors=iccvblue]{hyperref}

\def\paperID{1665} 
\def\confName{ICCV}
\def\confYear{2025}

\title{Future-Aware Interaction Network For Motion Forecasting}

\author{
    Shijie Li$^\clubsuit$\quad Xun Xu$^\clubsuit$\quad Si Yong Yeo $^\spadesuit$\quad Xulei Yang$^\clubsuit$
    \\[1ex]
    $\clubsuit$ I$^2$R, A*STAR~~~
    $\spadesuit$ Nanyang Technological University
 \\[1ex]
}

\begin{document}
\maketitle
\begin{abstract}
Motion forecasting is a crucial component of autonomous driving systems, enabling the generation of accurate and smooth future trajectories to ensure safe navigation to the destination. In previous methods, potential future trajectories are often absent in the scene encoding stage, which may lead to suboptimal outcomes. Additionally, prior approaches typically employ transformer architectures for spatiotemporal modeling of trajectories and map information, which suffer from the quadratic scaling complexity of the transformer architecture. In this work, we propose an interaction-based method, named Future-Aware Interaction Network, that introduces potential future trajectories into scene encoding for a comprehensive traffic representation. Furthermore, a State Space Model (SSM), specifically Mamba, is introduced for both spatial and temporal modeling. To adapt Mamba for spatial interaction modeling, we propose an adaptive reordering strategy that transforms unordered data into a structured sequence. Additionally, Mamba is employed to refine generated future trajectories temporally, ensuring more consistent predictions. These enhancements not only improve model efficiency but also enhance the accuracy and diversity of predictions.
We conduct comprehensive experiments on the widely used Argoverse 1 and Argoverse 2 datasets, demonstrating that the proposed method achieves superior performance compared to previous approaches in a more efficient way. The code will be released according to the acceptance.
\end{abstract}    
\section{Introduction} \label{sec:intro}
Motion forecasting is a critical component of autonomous driving systems. It serves as a bridge between the perception module and the action execution module by predicting accurate future trajectories, enabling safe and smooth navigation to the destination. 
However, the complexity of real-world traffic scenarios and the inherently multimodal nature of motion forecasting, where multiple plausible future trajectories can exist, still pose significant challenges to its practical application.

\begin{figure}[t]
    \centering
    \begin{minipage}{0.48\linewidth}
        \centering
        \includegraphics[width=\linewidth]{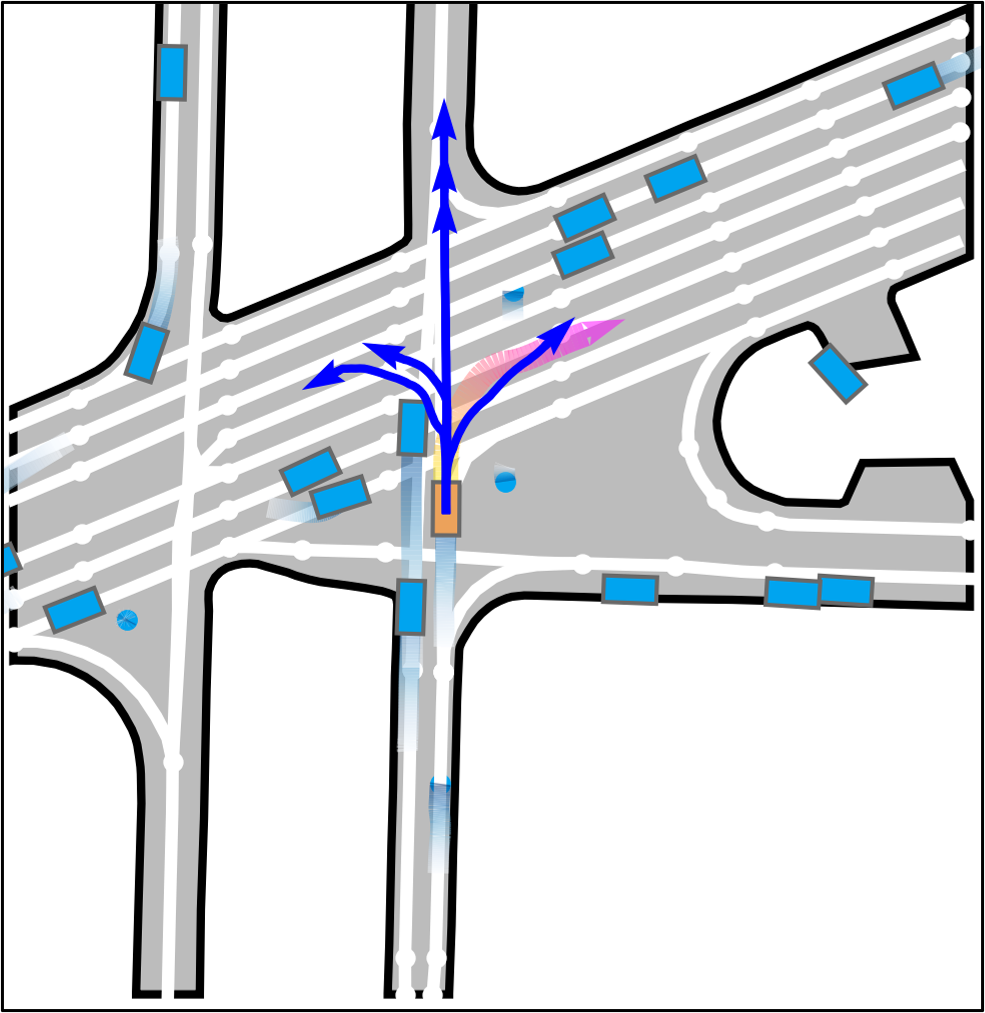}
        \subcaption{Query-based }
    \end{minipage}
    \hspace{1mm}
    \begin{minipage}{0.48\linewidth}
        \centering
        \includegraphics[width=\linewidth]{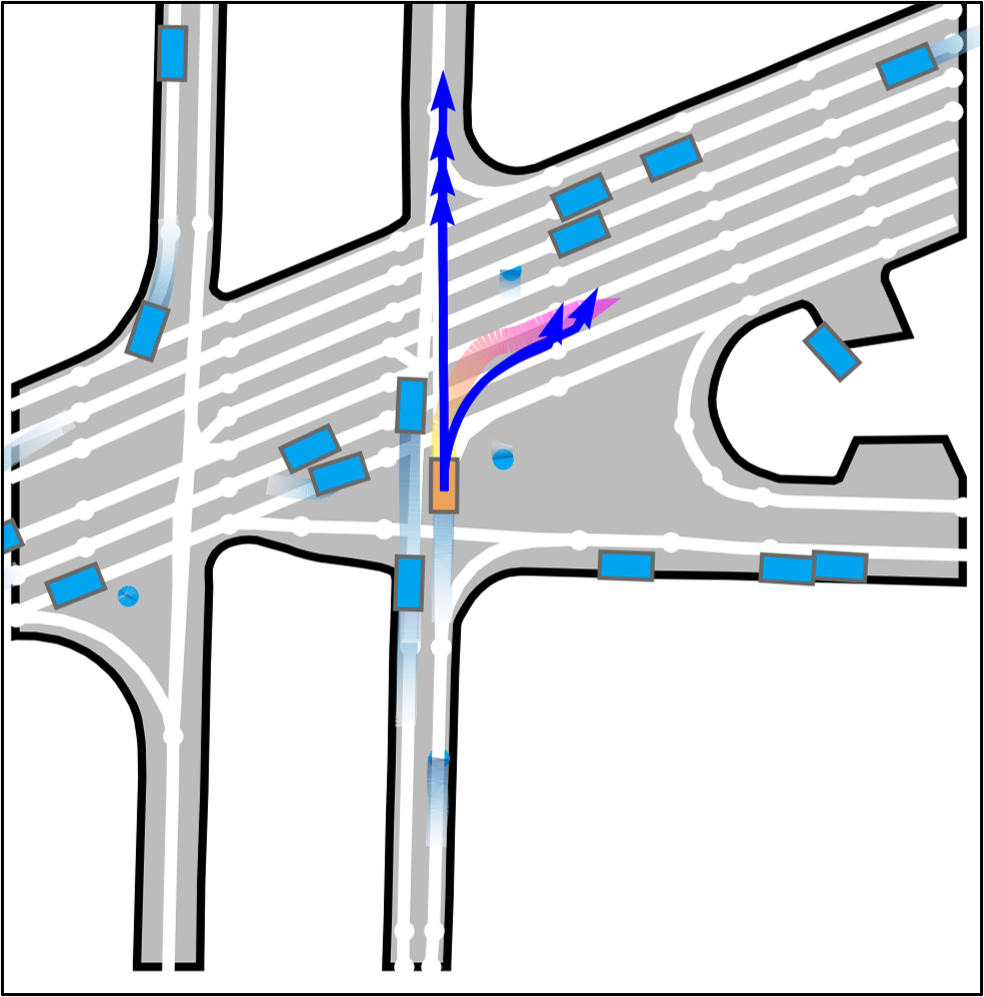}
        \subcaption{Intention-based }
    \end{minipage}
    \vspace{-3mm}
    \caption{ Query-based method (QCNet\cite{zhou2023query}) usually struggles to avoid unrealistic predictions (e.g., incorrectly predicting a left turn) due to the separate optimization of historical and future states. The proposed interaction-based method captures a more comprehensive representation of traffic scenarios by jointly optimizing the current state and future potential trajectories, leading to a deeper understanding and effectively reducing such errors. }
    \label{fig:showcase}
    \vspace{-6mm}
\end{figure}

\begin{figure*}[t]
    \centering
    \includegraphics[width=0.95\linewidth]{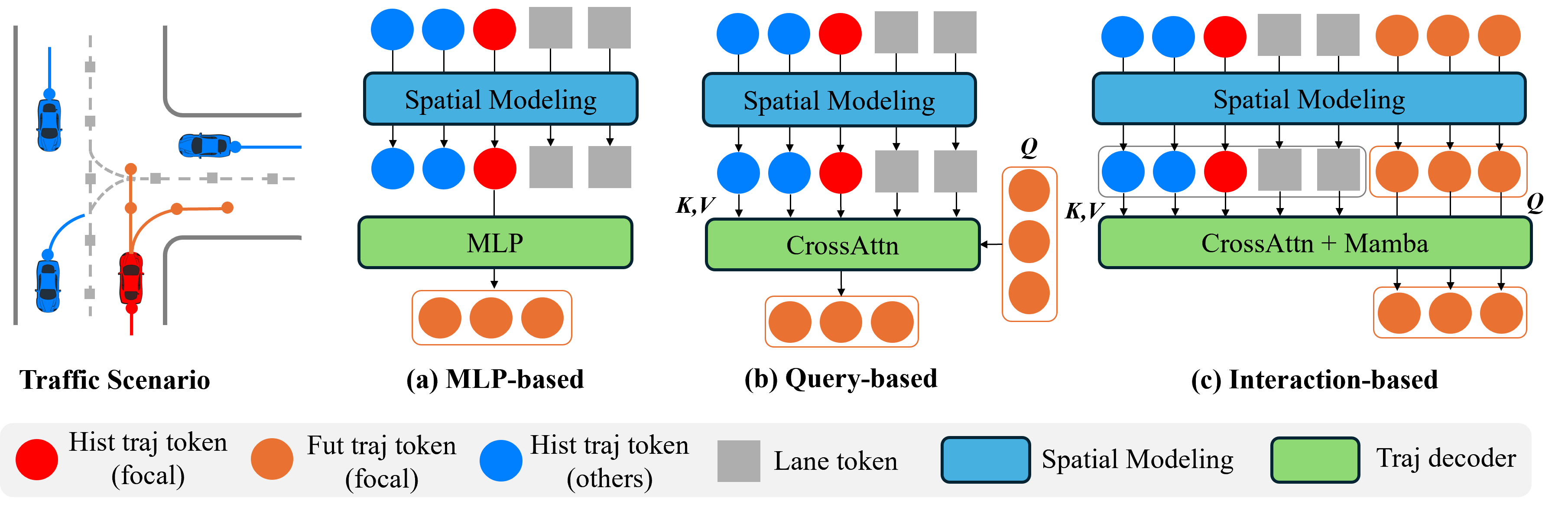}
     \vspace{-4mm}
    \caption{Previous methods can be categorized into \textbf{MLP-based} or \textbf{Query-based} approaches, where future trajectories are absent during the scene encoding stage. In contrast, we propose an \textbf{Interaction-based} method that models future trajectories in advance and seamlessly integrates them into scene encoding, enabling a more comprehensive representation.}
    \label{fig:intention_model}
    \vspace{-5mm}
\end{figure*}

Recent works \cite{zhou2023query,HiVT,Vectornet} typically encode the entire scenario into multiple tokens, with each token representing an individual scene element, agent's historical trajectory or a lane segment.
The future trajectories will either originate directly from the agent's current state with a Multilayer perception (MLP) or be represented as several learnable queries that denote driver intention. These queries aggregate relevant information from the encoded representations and are ultimately used to decode the future trajectories. We categorize these approaches as \textbf{MLP-based}\cite{HiVT} and \textbf{Query-based}\cite{zhou2023query} methods, respectively as indicated in Fig. \ref{fig:intention_model}. We observe these approaches usually introduce future trajectories after the scene encoding stage which results in the separate optimization of historical and future states, which may be suboptimal, as indicated in Fig. \ref{fig:showcase}.



The Transformer \cite{vaswani2017attention} is widely adopted for spatiotemporal modeling in motion forecasting tasks; however, its computational complexity scales quadratically with sequence length. In scenarios where numerous agents coexist, this creates significant challenges in maintaining efficiency while ensuring safety in such complex environments. Although the recently proposed Mamba architecture \cite{Mamba} exhibits remarkable efficiency, it is specifically tailored for sequential data processing and cannot be directly adapted to model spatial relationships.

In this work, we introduce a highly efficient architecture for motion forecasting, called the Future-Aware Interaction Network (\textbf{FINet}). FINet stands out from previous methods in two key ways. First, it models future states in advance and integrates them into the scene encoding stage. This enables the joint optimization of both historical and future states, allowing all scene elements to be aware of potential future information. As a result, a comprehensive traffic representation is learned. We refer to this strategy as the \textbf{Interaction-based} approach, as future trajectories actively participate in spatial interactions. Additionally, FINet uses Mamba for spatio-temporal modeling, significantly improving efficiency and ensuring consistent trajectory generation.

To integrate potential future trajectories that do not yet exist into scene encoding, a representative and expressive trajectory representation is essential. As mentioned earlier, future trajectories are influenced by both current motion states and driver intention. To account for both factors, we propose a strategy that effectively incorporates them. Additionally, we introduce a relevant inductive bias to further enhance the representation of future trajectories. By introducing future trajectories into scene encoding, FINet fosters mutual awareness among all traffic participants and facilitates their joint optimization, which contributes to a comprehensive traffic representation and prevents unrealistic predictions, as indicated in Fig. \ref{fig:showcase}. 
Furthermore, interactions among future trajectories are also considered, leading to more diverse predictions and a broader coverage of plausible future scenarios.

Mamba is integrated into FINet for spatiotemporal modeling, enhancing efficiency and ensuring consistent future trajectory generation. Since Mamba is specifically designed for sequential data modeling, it cannot be directly applied to spatial interactions, where data lacks an inherent sequential order. To address this limitation, we propose the Adaptive Reorder Strategy (ARS), which dynamically transforms unordered data into an ordered format in a learnable manner, enabling Mamba’s effective use for spatial interactions. Additionally, Mamba is introduced into the trajectory generation process. By extending future trajectories along the time dimension, Mamba refines them temporally, ensuring more consistent and coherent predictions.

The proposed method demonstrates superior performance while maintaining high efficiency on widely used motion forecasting datasets, including Argoverse2 and Argoverse1. Its low latency and minimal GPU memory usage make it well-suited for real-world applications. The main contributions of this paper can be summarized as:
\begin{itemize}
\item We propose the Future-Aware Interaction Network (FINet), which incorporates future trajectories into scene encoding. By enabling mutual awareness among all elements, FINet ensures accurate while consistent future trajectories prediction.
\item FINet incorporates Mamba for spatiotemporal modeling, further enhancing model efficiency while ensuring more consistent predictions.
\item  The proposed method demonstrates superior performance on the Argoverse2 and Argoverse1 motion forecasting datasets, achieving significantly improved efficiency while maintaining high accuracy.
\end{itemize}

\section{Related Work} \label{sec:related_work}

\subsection{Motion Forecasting}

Motion forecasting is a crucial component of autonomous driving systems, typically formulated as a spatiotemporal modeling task. It requires capturing temporal dependencies in historical trajectories, as well as spatial relationships among different agents and map information. 
To capture temporal information, recurrent neural networks (RNNs) \citep{RNN_1,Social-gan,Social-LSTM,Trajectron++,RNN_2} are widely employed due to their strong capability in processing sequential data. To incorporate spatial modeling, RNNs are often combined with convolutional neural networks (CNNs) \citep{Spatial-Temporal1,Spatial-Temporal2,Spatial-Temporal3,Spatial-Temporal4,RNN_2}, which are used to process rasterized map representations \citep{Rasterized_1,Rasterized_2,Rasterized_3,Rasterized_4}.
However, raster map representations often suffer from a loss of fine-grained scene details and inefficient processing. Similarly, RNNs handle trajectory data with limited efficiency, making them less suitable for meeting the high-efficiency requirements of autonomous driving systems.
Recently, a more compact vectorized representation of maps has been introduced \citep{GNN_1,Vectornet,gu2021densetnt,Lanercnn,Tnt}. This approach retains only the essential information needed for motion forecasting, representing it in a trajectory-like manner. Based on it, transformer architecture \cite{vaswani2017attention} is applied to process both temporal and spatial modeling \citep{LaneGCN, liu2021multimodal, HiVT}. 
However, in transformers, dependencies are implicitly incorporated through positional encoding, which may struggle to capture fine-grained relationships explicitly. Moreover, its computational complexity scales quadratically with the sequence length, making it inefficient for handling long sequences—a common scenario in autonomous driving, where numerous agents and interactions need to be modeled. 
In this work, we introduce the State Space Model, specifically Mamba, to enhance model efficiency and improve the representation of spatial information.

\subsection{State Space Models}
State Space Models (SSMs) are essential for modeling dynamic systems, leveraging latent variables to capture temporal dynamics. While widely used in areas like reinforcement learning \citep{Reinforcement_Learning} and linear dynamical systems \citep{Linear_Dynamical_Syatems}, SSMs face challenges in efficiently modeling long-range dependencies. Recent advancements, such as the Structured State Space Sequence model (S4) \citep{S4}, improve computational efficiency, and further innovations like H3 \citep{HHH} and Gated State Space (GSS) \citep{GSS} enhance modeling capabilities and hardware efficiency.
Mamba \citep{Mamba}, a recent SSM, has gained attention for its input selection mechanism and hardware-aware parallelism, enabling linear complexity and excellent performance in natural language processing \citep{GraphMamba,Mamba4Rec,C_Mamba} and computer vision \citep{Vision_mamba,Mambair,VideoMamba,PointMamba,VideoMambaPro}. However, its application to motion forecasting remains unexplored. In this work, we integrate Mamba into motion forecasting for spatiotemporal modeling, enhancing the model's capacity and efficiency.

\section{Preliminaries}

\begin{figure*}[t]
    \centering
    \includegraphics[width=0.96\linewidth]{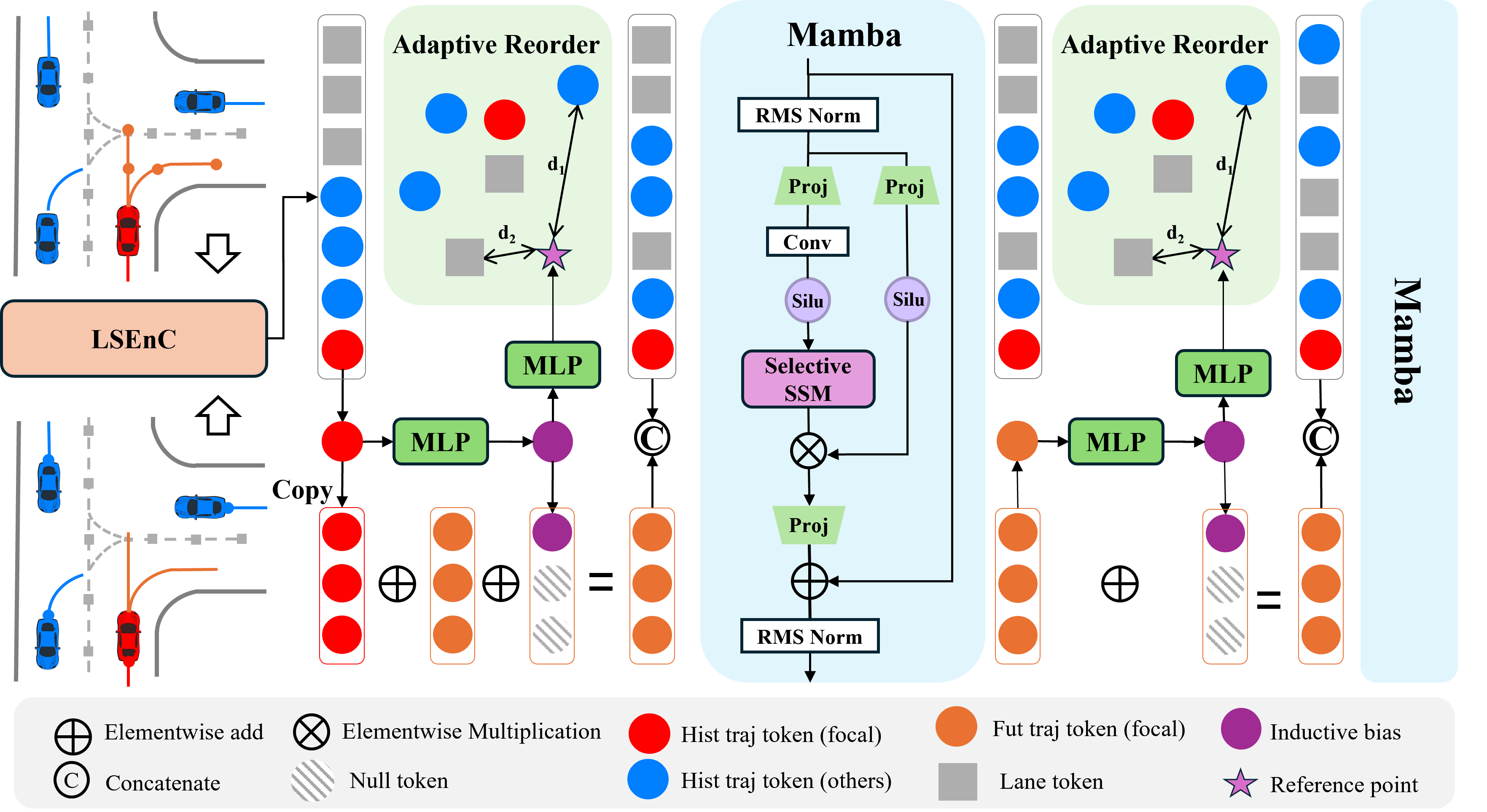}
    \vspace{-3mm}
    \caption{The proposed Future-Aware Interaction Mamba (FIM). The future potential trajectories will be modeled in advance and then integrated into the scene encoding. By enabling the model to be aware of future states, a more comprehensive representation can be learned.}
    \label{fig:fim}
    \vspace{-5mm}
\end{figure*}

 We begin by outlining the necessary preliminaries, including the task definition for motion forecasting and a brief overview of the State Space Model \citep{Mamba}.

\subsection{Motion Forecasting}
\label{sec:motion_forecasting}
Motion forecasting in autonomous driving scenarios is typically formulated as a spatiotemporal modeling task, where the goal is to predict multiple potential future trajectories as outputs. The inputs usually include both agent historical trajectories $\mathcal{T}^{hist}$ and the surrounding lane map $\mathcal{LM}$: 
\begin{equation}
\mathcal{T}^{hist} = \{x_t : t \in \{-T_h+1, \ldots , 0\}\}_{0:N_a}.
\end{equation}
\begin{equation}
\mathcal{LM} = \{ x_i : i \in \{1, N_{pt}\}\}_{1:N_l}
\end{equation}
We assume there are $N_a + 1$ agents in the same scenario, where the $0$-th agent represents the focal agent whose future trajectory is to be predicted. The historical trajectory of each agent is represented as a sequence of 2D points $x_t$, spanning from the starting timestamp $t = -T_h$ to the current timestamp $t = 0$. Additionally, the lane map $\mathcal{LM}$ consists of $N_l$ lane segments, each represented by $N_{pt}$ uniformly sampled 2D points along its centerline.

In motion forecasting, $K$ potential future trajectories, along with their corresponding probability scores, are typically predicted to address the ambiguity arising from the existence of multiple plausible future trajectories. This can be formulated as:
\begin{equation}
    (\mathcal{\hat{T}}^{fut}_k, \hat{s}_k)_{1:K} = \mathbf{Model}(\mathcal{T}^{hist}, \mathcal{LM}).
\end{equation}
where $\hat{s}_k \in (0, 1)$ is the probability score for $k-th$ predicted trajectory $\mathcal{\hat{T}}^{fut}_k$: 
\begin{equation}
\mathcal{\hat{T}}^{fut}_{k} = \{\hat{x}_t : t \in \{1, \ldots, T_f\}\}_{k}
\end{equation}
where $\hat{x}_t$ is the predicted 2D position at timestamp $t$. $T_f$ is the number of future steps that need to be predicted.

Correspondingly, the ground-truth future trajectory of the focal agent is defined as:   
\begin{equation}
\mathcal{T}^{fut} = \{x_t : t \in \{1, \ldots, T_f\}\},
\end{equation}

\subsection{Selective State Space Model}

State Space Models (SSMs), such as Structured State Space Sequence Models (S4) and Mamba, excel at processing long sequences by mapping a one-dimensional input $x(t) \in \mathbb{R}$ to an output $y(t) \in \mathbb{R}$ via a hidden state $h(t) \in \mathbb{R}^N$. The system is governed by matrices $A \in \mathbb{R}^{N \times N}$, $B \in \mathbb{R}^{N \times 1}$, and $C \in \mathbb{R}^{1 \times N}$ for state transitions, input projections, and output projections, respectively.

The discretized system, with step size $\Delta$, is defined as:
\begin{equation}
    h_t = A h_{t-1} + B x_t,
\end{equation}
\begin{equation}
    y_t = C h_t.
\end{equation}

Output computation utilizes a structured convolutional kernel $K$ spanning sequence length $M$:
\begin{equation}
    K = (CB, CAB, \dots, CA^{M-1}B),
\end{equation}
\begin{equation}
    y = x \ast K,
\end{equation}
where $\ast$ denotes the convolution operation.
\section{Methodology}
\label{sec:method}

In this section, we introduce the proposed Future-Aware Interaction Network (FINet) shown in Fig. \ref{fig:fim}, which comprises three main components: a lightweight scene encoder that converts the input scene into multiple tokens (Sec. \ref{sec:scene_encoder}); the Future-Aware Mamba, which models and incorporates potential future trajectories into scene encoding (Sec. \ref{sec:fi_mamba}); and a Temporal Enhanced Decoder (TEDec) that decodes future trajectories (Sec. \ref{sec:dec}). Finally, we detail the applied supervision strategies in Sec. \ref{sec:loss}.

\subsection{Lightweight Scene Encoder \textbf{(LSEnc)}}
\label{sec:scene_encoder}
We propose a lightweight scene encoder that converts input scenes into a token-based representation, mapping each trajectory or lane segment to an individual token efficiently.

The agent's historical trajectory is encoded using a stack of Mamba blocks ($\mathbf{MambaBlocks}$), which process each historical trajectory $\mathcal{T}^{hist}_i$ sequentially from the beginning to the end with linear complexity. The final token ($t=0$) is then selected to represent the entire trajectory:
\begin{equation}
    \mathcal{ST}_i^A = \mathbf{MambaBlocks}(\mathcal{T}^{hist}_i)[0] + Cls^{A}_i,
\end{equation}
$Cls^{A}_i$ denotes the semantic information of agents, such as vehicles or pedestrians.

For the lane map $\mathcal{LM}$, which typically contains significantly more points compared to trajectory data, we employ a lightweight mini-PointNet \citep{PointNet} to efficiently learn lane embeddings. For each lane segment $\mathcal{L}_i$:
\begin{equation}
     \mathcal{ST}_i^L = \mathbf{MiniPointNet}(\mathcal{L}_i) + Cls^{L}_i,
\end{equation}
Similarly, $Cls^{L}_i$ represents the lane types and is initialized as a learnable embedding. 

By separately encoding each trajectory and lane segment, the entire scene is transformed into a token-based representation $\mathcal{ST}$. The encoded scene tokens include trajectory tokens ($\mathcal{ST}^A = \{ \mathcal{ST}^A_i \mid i \in {0, ..., N_a}\}$) and lane tokens ($\mathcal{ST}^L = \{ \mathcal{ST}^L_i \mid i \in \{0, ..., N_l\}\}$):
\begin{equation}
    \mathcal{ST} = (\mathcal{ST}^A || \mathcal{ST}^L)
    \label{eq:scene_token}
\end{equation}
Here, $||$ denotes the concatenation operator. This design achieves a balance between model efficiency and performance by leveraging the Mamba architecture and strategically allocating computational resources.

\begin{figure}[t]
    \centering
    \includegraphics[width=0.95\linewidth]{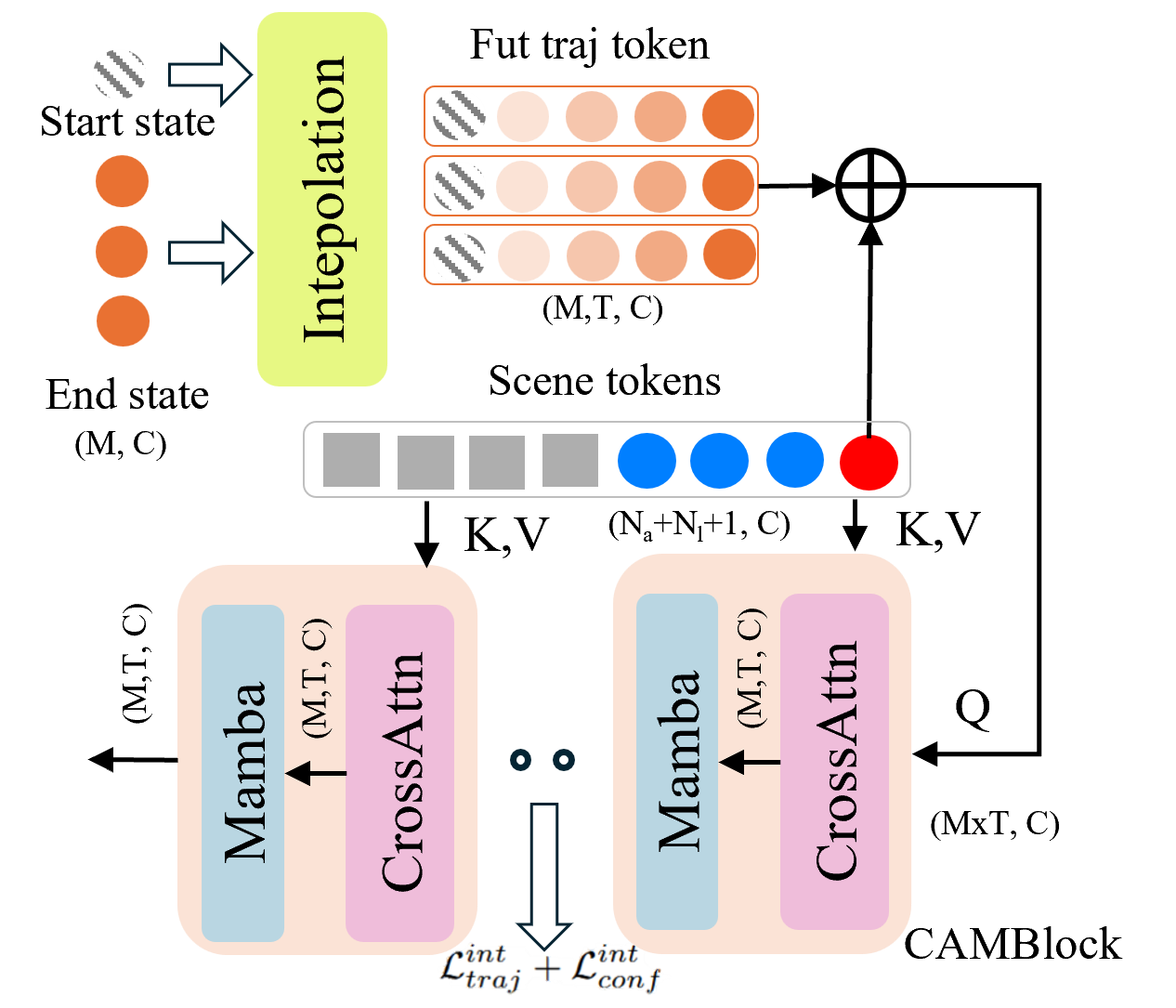}
    \vspace{-4mm}
    \caption{The proposed Temporal Enhanced Decoder (TEDec). For simplicity, we omit reshape operation but indicate the tensor shape at each step.}
    \label{fig:dec}
    \vspace{-5mm}
\end{figure}

\subsection{Future-Aware Interaction Mamba \textbf{(FIM)}} \label{sec:fi_mamba}
The Future-Aware Interaction Mamba (FIM), as shown in Fig. \ref{fig:fim}, models future potential trajectories in advance and jointly optimizes them with current scene elements spatially. By enabling the model to be aware of future states, it learns a more comprehensive representation, contributing to both accurate and diverse predictions. FIM is achieved through the Mamba architecture, which enhances performance and efficiency, as demonstrated by experimental results.


\textbf{Future Trajectories Modeling} The first challenge is how to model potential future trajectories, which is even absent in annotation. We observe that future trajectories are primarily influenced by the current motion state, represented by $\mathcal{T}^{hist}_0 \in \mathbb{R}^{1 \times C}$, which is equivalent to $\mathcal{ST}^A_0$, and by driver intentions, which are modeled using $K$ learnable tokens, $\mathcal{T}^{DI} \in \mathbb{R}^{K \times C}$. Additionally, we introduce a relevant inductive bias $\mathcal{T}^{bias} \in \mathcal{R}^{1\times C}$, which will be discussed in detail later.
The potential future trajectories are:
\begin{equation}
    \mathcal{T}^{fut} = \mathcal{T}^{hist}_0 + \mathcal{T}^{bias} + \mathcal{T}^{DI}, \mathcal{T}^{bias} \in \mathcal{R}^{K\times C}
    \label{eq:fut_traj_model}
\end{equation}
The $\mathcal{T}^{bias}$ is added only to the first trajectory and then propagated to other future 
 potential trajectories through the Mamba block, based on experimental results. We hypothesize that this approach facilitates a globally optimal distribution of potential future trajectories.

The Transformer \cite{vaswani2017attention} is widely used for spatial interaction. However, its computational complexity scales quadratically with sequence length, posing a challenge in autonomous driving scenarios where numerous agents and interactions exist. The Mamba architecture offers greater efficiency, scaling linearly with sequence length. However, it is inherently designed for sequential data processing and cannot be directly applied to spatial interactions, where scene elements are typically unordered.

\textbf{Adaptive Reorder Strategy (ARS)} To convert unordered scene elements into an ordered sequence, it is necessary to define a specific order. Since no ground truth order exists, we propose an Adaptive Reorder Strategy that determines the scene element order in a learnable manner. Specifically, we predict an offset from center to a learnable reference point. All elements are then sorted by their distance to the reference point, forming an ordered sequence.

In FIM, the scene tokens $\mathcal{ST}$ are first reordered using the Adaptive Reorder Strategy (ARS), where the reference point $RP$ is predicted by $\mathcal{T}^{hist}_0$ through two Multilayer Perceptron ($\mathbf{MLP}$):
\begin{equation}
    \mathcal{T}^{bias}_1 = \mathbf{MLP}_{1}^{b}(\mathcal{T}^{hist}_0)
\end{equation}
\begin{equation}
    \mathcal{RP}_1 = \mathbf{MLP}_{1}^{o}(\mathcal{T}^{bias}_1)
\end{equation}
Here, we replace the predicted offset with the predicted reference point $\mathcal{RP}_1$ for simplicity. $\mathcal{T}^{bias}_1$ refers to the inductive bias as defined in Eq. \ref{eq:fut_traj_model}. The scene tokens are then reordered according to their spatial distance to $\mathcal{RP}_1$:
\begin{equation}
    \mathcal{OST}_1 = \mathbf{Reorder}(\mathcal{ST}, \mathcal{RP}_1).
\end{equation}
One important detail to note is that the token corresponding to the focal agent is placed at the end, ensuring it has the greatest influence on the future trajectory tokens $\mathcal{T}^{fut}$, which are concatenated at the end:
\begin{equation}
    \mathcal{OSTF}_1 = (\mathcal{OST}_1 || \mathcal{T}^{fut})
\end{equation}
A stack of bidirectional Mamba blocks is applied to model the spatial interaction among all tokens:
\begin{equation}
    \mathcal{OSTF}'_1 = \mathbf{BiMambaBlocks}(\mathcal{OSTF}_1)
\end{equation}






The above procedure will be repeated, with the key difference that the reference point $\mathcal{RP}_2$ is predicted by the first future trajectory token, $\mathcal{T}^{fut}_0$, from the previous stage. In addition, auxiliary supervision is applied to align the predicted reference point $\mathcal{RP}_2$ with the end point of the ground truth future trajectory.  In this way, the tokens closer to the predicted endpoints will have a greater impact, enhancing the outcomes.


In previous methods, future trajectories depend on the scene elements, $P(\mathcal{\hat{T}}^{fut} | \mathcal{ST})$. In FIM, they are modeled jointly as $P(\mathcal{\hat{T}}^{fut}, \mathcal{ST})$. This enables mutual awareness between scene elements and future trajectories, fostering a more comprehensive traffic representation through deep interactions. Additionally, each future trajectory is aware of other predicted future trajectories, promoting more diverse predictions and better coverage of a broader range of potential outcomes.

\begin{table*}[t]
\centering
\setlength{\tabcolsep}{1mm}
\resizebox{0.9\textwidth}{!}{
\begin{tabular}{c|cccc|ccc}
\toprule
\textbf{Method} & \textbf{b-minFDE$_6$} ($\downarrow$) & \textbf{minADE$_6$} ($\downarrow$) & \textbf{minFDE$_6$} ($\downarrow$) & \textbf{MR$_6$} ($\downarrow$) & \textbf{minADE$_1$} ($\downarrow$) & \textbf{minFDE$_1$} ($\downarrow$) & \textbf{MR$_1$} ($\downarrow$) \\ 
\midrule
GoRela\citep{cui2023gorela}      & 2.01               & 0.76             & 1.48             & 0.22          & 1.82              & 4.62             & 0.66          \\ 
THOMAS\citep{gilles2022thomas}     & 2.16               & 0.88        
& 1.51             & 0.20          & 1.95              & 4.71             & 0.64          \\ 
MTR \citep{shi2022motion}        & 1.98               & 0.73             & 1.44             & 0.15          & 1.74              & 4.39             & \textbf{0.58}         \\ 
GANet \citep{wang2023ganet}     & 1.96               & 0.72             & 1.34             & 0.17          & 1.77              & 4.48             & \underline{0.59}          \\ 
MacFormer \citep{feng2023macformer}       & 1.91  & 0.70  & 1.38  & 0.19  & 1.84 & 4.69  & 0.61           \\ 
GNet\citep{gao2023dynamic}            & \underline{1.90}  & 0.69  & 1.34  & 0.18  & 1.72 & 4.40  & \underline{0.59}           \\ 
ProphNet \citep{wang2023prophnet}       & \textbf{1.88}  & \underline{0.66}  & 1.32  & 0.18  & 1.76 & 4.77  & 0.61           \\ 
Forecast-MAE \cite{Forecast-MAE} & 2.03  & 0.71  & 1.39   & 0.17   & 1.74  & 4.36  & 0.61  \\
FRM\cite{park2023leveraging} & 2.47                  & 0.89                & 1.81                & 0.29            & 2.37                & 5.93               & 0.89 \\
HDGT\cite{jia2023hdgt} &  2.24                  & 0.84                & 1.60                & 0.21            & 2.08                & 5.37               & 0.84           \\
SIMPL\cite{zhang2024simpl} & 2.05                  & 0.72                & 1.43                & 0.19            & 2.03                & 5.50               & 0.76     \\
QCNet \citep{zhou2023query}  & 1.91         & \textbf{0.65}             & \underline{1.29}             & \underline{0.16}          & \underline{1.69}              & \underline{4.30}             & \underline{0.59}          \\ 
\midrule
FINet (Ours)  & 1.93 & \underline{0.66} & \textbf{1.27} & \textbf{0.15} & \textbf{1.60} & \textbf{4.02} & \textbf{0.57} \\ 
  
\bottomrule
\end{tabular}
}
\caption{Comparison of motion forecasting methods on the Argoverse 2 test set. For each metric, the best result is highlighted in  \textbf{bold} while the second-best result is \underline{underlined}.}
\label{tab:sota_comparison_av2}
\vspace{-5mm}
\end{table*}

\subsection{Temporal Enhanced Decoder \textbf{(TEDec)}}  \label{sec:dec}
Finally, the Temporal Enhanced Decoder (\textbf{TEDec}), shown in Fig. \ref{fig:dec}, will decode the future trajectories.
It receives tokens from FIM, which are categorized into scene tokens $\mathcal{ST}^{scene} \in \mathbb{R}^{(N_a + N_l + 1) \times C}$, representing the current scenario, and future trajectory tokens $\mathcal{ST}^{fut} \in \mathbb{R}^{K \times C}$, corresponding to $K$ predicted future trajectories.

In previous methods, each future trajectory is predicted from a single token $\mathcal{ST}^{fut}_i$. This strategy is suboptimal as it cannot ensure temporal consistency of predicted future trajectory.
In TEDec, $\mathcal{ST}^{fut}_i$ is first extended into a sequential format. Then, it aggregates relevant information from scene tokens $\mathcal{ST}^{scene}$ and is temporally refined by the Mamba block to ensure its temporal consistency.
As the future trajectory starts from the historical motion state $\mathcal{ST}^{scene}_0$ corresponding to focal agent and gradually evolves into $\mathcal{ST}^{scene}_0 + \mathcal{ST}^{fut}$. The future trajectory will be derived through interpolation between them:
\begin{equation}
    \mathcal{IDT}^{fut} = \frac{t}{T^{fut}} \cdot \mathcal{ST}^{fut}, \quad t=0, ..., T_f
\end{equation}
where $\mathcal{IDT}^{fut} \in \mathcal{R}^{K \times T^{fut} \times C}$, Then the interpolated tokens $\mathcal{IDT}^{fut}$ will be added to  $\mathcal{ST}^{scene}_0$:
\begin{equation}
    \mathcal{DT}^{fut} = \mathcal{ST}^{scene}_0 + \mathcal{IDT}^{fut}
\end{equation}

Thus, the resulting tokens $\mathcal{DT}^{fut}$ will span from the current token ($\mathcal{ST}^{scene}_0$) to the future token ($\mathcal{ST}^{scene}_0 + \mathcal{ST}^{fut}$), providing a more comprehensive representation of the changing motion states.

$\mathcal{DT}^{fut}$ is then processed through several \textbf{CAMBlock}, where each block consists of a cross-attention layer ($\mathbf{CA}$) followed by a bidirectional Mamba block ($\mathbf{M}$). This process can be formulated as:
\begin{equation}
    \mathcal{DT}^{fut}_{ca} = \mathbf{CA}(\mathcal{DT}^{fut}, \mathcal{ST})
\end{equation}
\begin{equation}
    \mathcal{DT}^{fut}_m = \mathbf{M}(\mathcal{DT}^{fut}_{ca})
\end{equation}
In each \textbf{CAMBlock}, tokens in future trajectories first aggregate the most relevant information from scene tokens $\mathcal{ST}$ using the cross-attention block. Subsequently, Mamba sequentially processes each trajectory from start to end, optimizing it to strengthen the temporal dependencies.

Finally, the future trajectories and corresponding scores are output through two Multilayer perceptron:
\begin{equation}
    \hat{\mathcal{T}}^{fut} = \mathbf{MLP}_{traj}(\mathcal{DT}^{fut})
\end{equation}
\begin{equation}
    \hat{s}^{fut} =\mathbf{MLP}_{score}(\mathbf{Maxpool}^{T}(\mathcal{DT}^{fut}))
\end{equation}
where $\mathbf{Maxpool}^{T}$ refers to applying max pooling along the temporal dimension.

\subsection{Supervision} \label{sec:loss}
For the final outputs, we employ the widely used smooth L1 loss for trajectory regression, $\mathcal{L}_{traj}$, and cross-entropy loss for confidence classification, $\mathcal{L}_{score}$. Additionally, we adopt the winner-take-all strategy, which optimizes only the best prediction—specifically, the one with the minimal final prediction error relative to the ground truth. To encourage consistent predictions throughout the decoding process, the same losses are applied to the intermediate outputs in TEDec, represented as $\mathcal{L}_{traj}^{int} + \mathcal{L}_{score}^{int}$. Furthermore, alignment between the second predicted offset, $\mathcal{RP}_2$, and the endpoint of the ground truth future trajectory, as detailed in Sec. \ref{sec:fi_mamba}, is enforced using a smooth L1 loss, $\mathcal{L}_{align}$. The overall loss is defined as:
\begin{equation}
    \mathcal{L} = \mathcal{L}_{traj} + \mathcal{L}_{score} + \mathcal{L}_{traj}^{int} + \mathcal{L}_{score}^{int} + L_{align}
\end{equation}

\begin{table}[ht]
\centering
\setlength{\tabcolsep}{1mm}
\resizebox{0.85\linewidth}{!}{
\begin{tabular}{lccc}
\hline
\textbf{Method} & \textbf{minADE$_6$} & \textbf{minFDE$_6$} & \textbf{MR$_6$} \\
\hline
LTP \cite{wang2022ltp} & 0.78 & 1.07 & - \\
LaneRCNN \cite{zeng2021lanercnn} & 0.77 & 1.19 & \textbf{0.08} \\
TPCN \cite{ye2021tpcn} & 0.73 & 1.15 & 0.11 \\
DenseTNT \cite{gu2021densetnt} & 0.73 & 1.05 & 0.10 \\
TNT \cite{Tnt} & 0.73 & 1.29 & 0.09 \\
mmTransformer \cite{liu2021multimodal} & 0.71 & 1.15 & 0.11 \\
LaneGCN \cite{LaneGCN} & 0.71 & 1.08 & - \\
SSL-Lanes \cite{SSL-Lane} & 0.70 & 1.01 & \underline{0.09} \\
PAGA \cite{da2022path} & 0.69 & 1.02 & \underline{0.09} \\
DSP \cite{zhang2022trajectory} & 0.69 & 0.98 & \underline{0.09} \\
FRM \cite{park2023leveraging} & 0.68 & 0.99 & - \\
ADAPT \cite{aydemir2023adapt} & 0.67 &  \textbf{0.95} & \textbf{0.08} \\
SIMPL \cite{zhang2024simpl} & \underline{0.66} &  \textbf{0.95} & \textbf{0.08} \\
HiVT \cite{HiVT} &\underline{ 0.66} & \underline{0.96} & \underline{0.09} \\
R-Pred \cite{choi2023r} & \underline{0.66} & \textbf{0.95} & \underline{0.09} \\
\midrule
FINet (Ours) &  \textbf{0.59} & \textbf{0.95} & \underline{0.09} \\
\hline
\end{tabular}}
\caption{Comparison of motion forecasting methods on the Argoverse 1 validation set. For each metric, the best result is highlighted in  \textbf{bold} while the second-best result is \underline{underlined}.}
\label{tab:sota_comparison_av1}
\vspace{-5mm}
\end{table}

\section{Experiments} \label{sec:exp}

\subsection{Experiment Setting}

\textbf{Dataset}
For evaluation, we use two widely adopted motion forecasting datasets: Argoverse2\cite{wilson2023argoverse} and Argoverse1\cite{chang2019argoverse}. Argoverse2 consists of 199,908 training sequences, 24,988 validation sequences, and 24,984 testing sequences. Each sequence is sampled at 10 Hz, containing 5 seconds of historical data and requiring predictions for the next 6 seconds ($T_h = 50$, $T_f = 60$). In comparison, Argoverse1 includes 205,942 training samples and 39,472 validation samples, with a total sequence duration of 5 seconds, comprising 2 seconds of observation and 3 seconds of prediction ($T_h = 20$, $T_f = 30$), also sampled at 10 Hz.

\textbf{Evaluation Metrics}
We evaluate our model using standard metrics for motion prediction: minimum Average Displacement Error (minADE), minimum Final Displacement Error (minFDE), and Miss Rate (MR). MinADE calculates the average distance $\ell_2$ between the best predicted and ground-truth trajectories in all time steps, while minFDE measures the distance $\ell_2$ in the final time step. MR is the proportion of cases where the best-predicted endpoint is more than 2.0 meters from the ground-truth endpoint. These metrics evaluate up to six predicted trajectories per agent, with the best trajectory having the smallest endpoint error.

\subsection{Comparison to previous methods}

We compare the proposed method with previous approaches on the Argoverse 2 dataset, and the experimental results are presented in Tab. \ref{tab:sota_comparison_av2}. The results indicate that the proposed method outperforms existing methods on almost all metrics, demonstrating its superior performance. Specifically, while ProphNet \cite{wang2023prophnet} achieves the best performance on b-minFDE$_6$, it performs significantly worse on other metrics. For instance, the proposed method surpasses ProphNet on minADE$_6$ by approximately 4\% (1.32 vs. 1.27) and on minFDE$_6$ by around 15\% (0.18 vs. 0.15). Furthermore, compared to QCNet \cite{zhou2023query}, which is a pure transformer-based method, the proposed method achieves comparable performance on minADE$_6$ (0.65 vs. 0.66) but outperforms it in almost all other metrics. In particular, the improvements in minADE$_1$ (5\%, 1.69 vs. 1.60) and minFDE$_1$ (6\%, 4.30 vs. 4.02) highlight the overall superior performance of the proposed method.
We observe that the performance improvement for $K=1$ is larger than that for $K=6$. This can be attributed not only to the larger improvement space for $K=1$, but also to the proposed method's ability to generate more diverse future trajectories and accurate scores by jointly optimizing both current and future states.

\begin{table}[t]
\centering
\resizebox{\linewidth}{!}{
\begin{tabular}{l|cccc}
\toprule
\textbf{Method} & \textbf{Flops (G)} & \textbf{Latency (ms)} & \textbf{Model size(M)} & GPU Mem (G) \\
\midrule
QCNet \cite{zhou2023query}     & 28.0   & 54.55     & 7.7 & 2.92\\
\midrule
FINet (Ours) & \textbf{1.47} &  \textbf{17.72} & \textbf{3.7} & \textbf{0.55} \\ 
\bottomrule
\end{tabular}}
\caption{Efficiency Analysis. The experiment was conducted on a single NVIDIA RTX A5000 with a batch size equals to 1.}
\label{tab:latency}
\vspace{-6mm}
\end{table}

\begin{table*}[t]
\centering
\setlength{\tabcolsep}{3mm}
\resizebox{\linewidth}{!}{
\begin{tabular}{c|cccc|ccc}
\toprule
\textbf{Method} & \textbf{b-minFDE$_6$} ($\downarrow$) & \textbf{minADE$_6$} ($\downarrow$) & \textbf{minFDE$_6$} ($\downarrow$) & \textbf{MR$_6$} ($\downarrow$) & \textbf{minADE$_1$} ($\downarrow$) & \textbf{minFDE$_1$} ($\downarrow$)  \\ 
\hline
MLP-based & 2.09 & 0.74 & 1.45 & 0.20 & 1.74 & 4.34 \\
Query-based & 2.08 & 0.73 & 1.43 & 0.18 & 1.73 & 4.28 \\
\hline
Interaction-based (w/o bias) & 1.99 & 0.66 & 1.32 & 0.16 & 1.60 & 4.03  \\
Interaction-based (all bias) & 1.98 & 0.66 & 1.35 & 0.16 & 1.60 & 4.02 \\
Interaction-based ($t=0$ bias) & 1.93 & 0.65 & 1.27 & 0.15 & 1.57 & 3.94 \\
\hline
\end{tabular}}
\vspace{-2mm}
\caption{Influenced of decoder type and inductive bias.}
\label{tab:ablation_decoder}
\vspace{-3mm}
\end{table*}

We then compare the proposed method with previous approaches on the Argoverse 1 dataset, with the experimental results presented in Tab. \ref{tab:sota_comparison_av1}. The results demonstrate that the proposed method achieves the best performance on minADE$_6$ and minFDE$_6$ while attaining the second-best performance on MR$_6$. Specifically, it outperforms previous methods on minADE$_6$, achieving an approximate 10\% improvement (0.66 vs. 0.59). These findings further demonstrate the superior effectiveness of the proposed method.

\begin{figure*}[htbp]
    \centering

    \subfloat[QCNet\cite{zhou2023query}]{
    \includegraphics[width=0.24\textwidth]{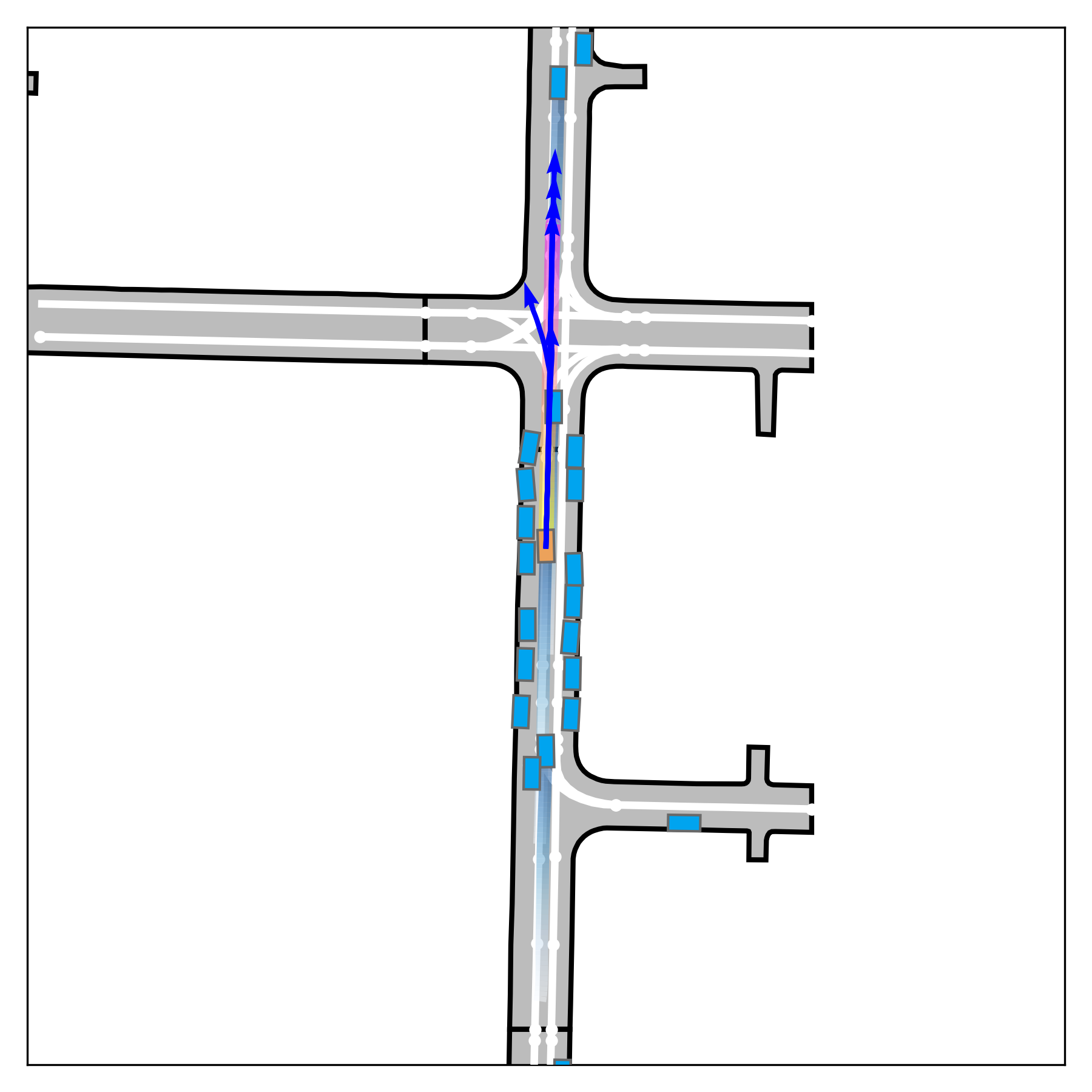} \hspace{-2.5mm}
    \includegraphics[width=0.24\textwidth]{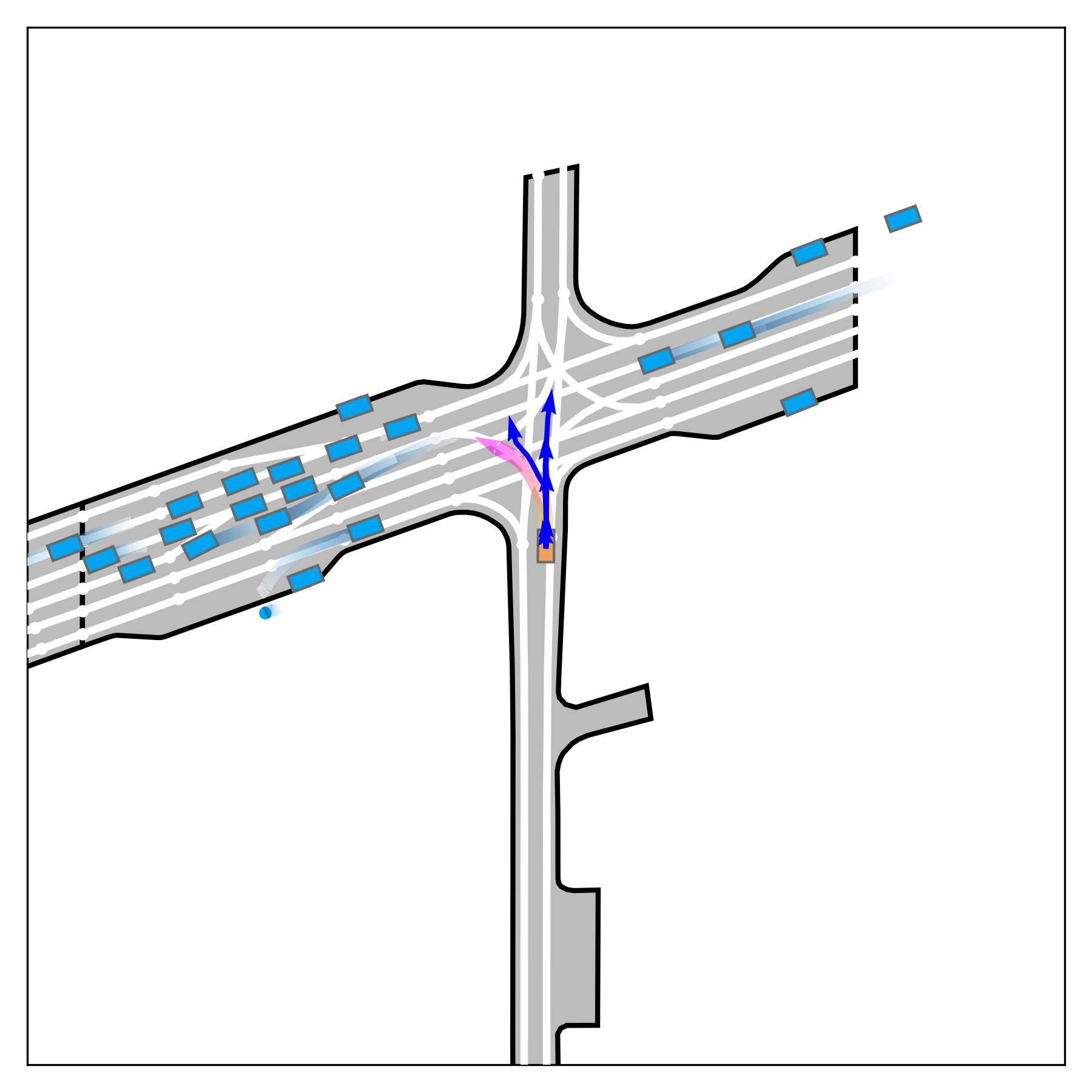} \hspace{-2.5mm}
    \includegraphics[width=0.24\textwidth]{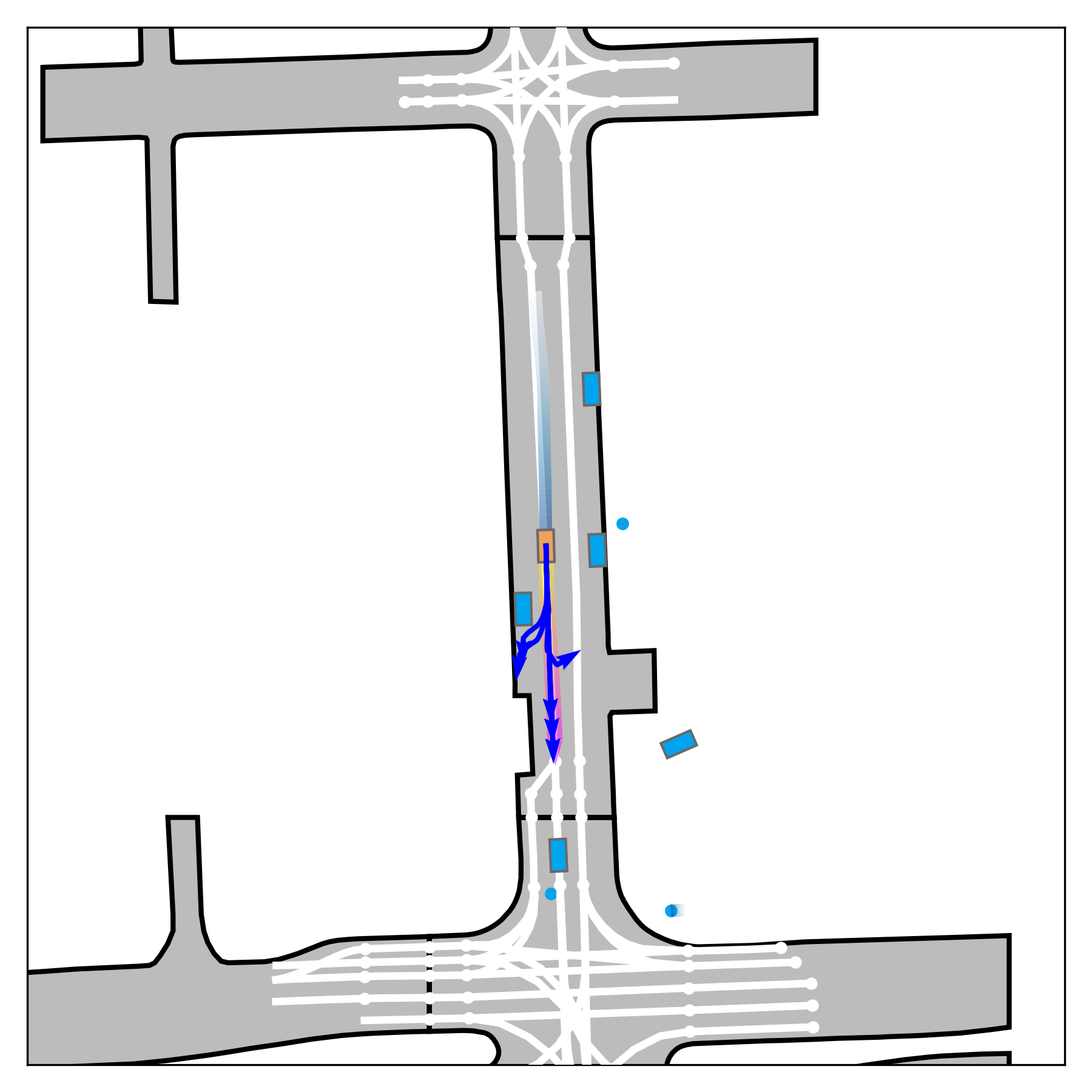} \hspace{-2.5mm}
    \includegraphics[width=0.24\textwidth]{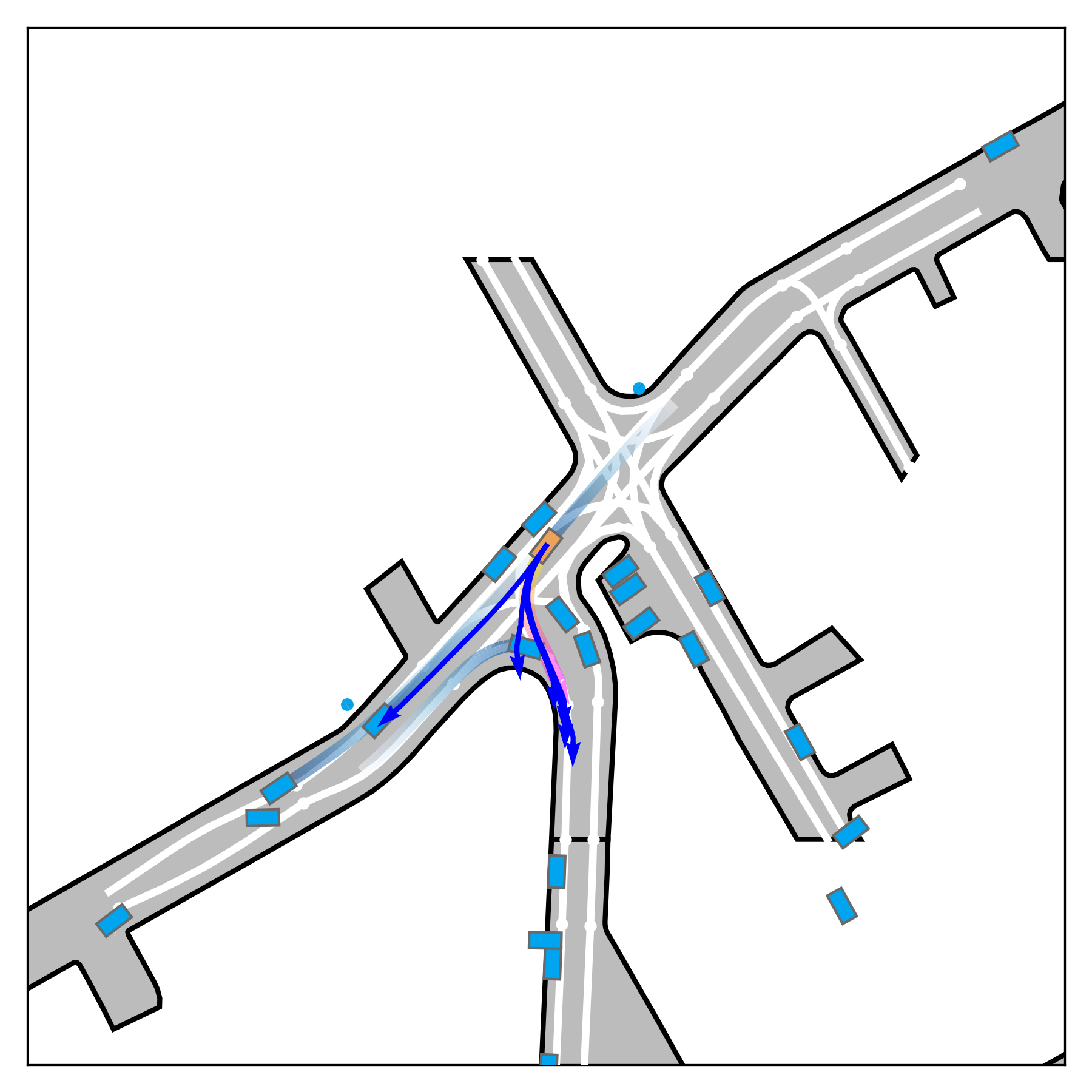} \hspace{-2.5mm}
    }
    
    \subfloat[FINet (Ours)]{
    \includegraphics[width=0.24\textwidth]{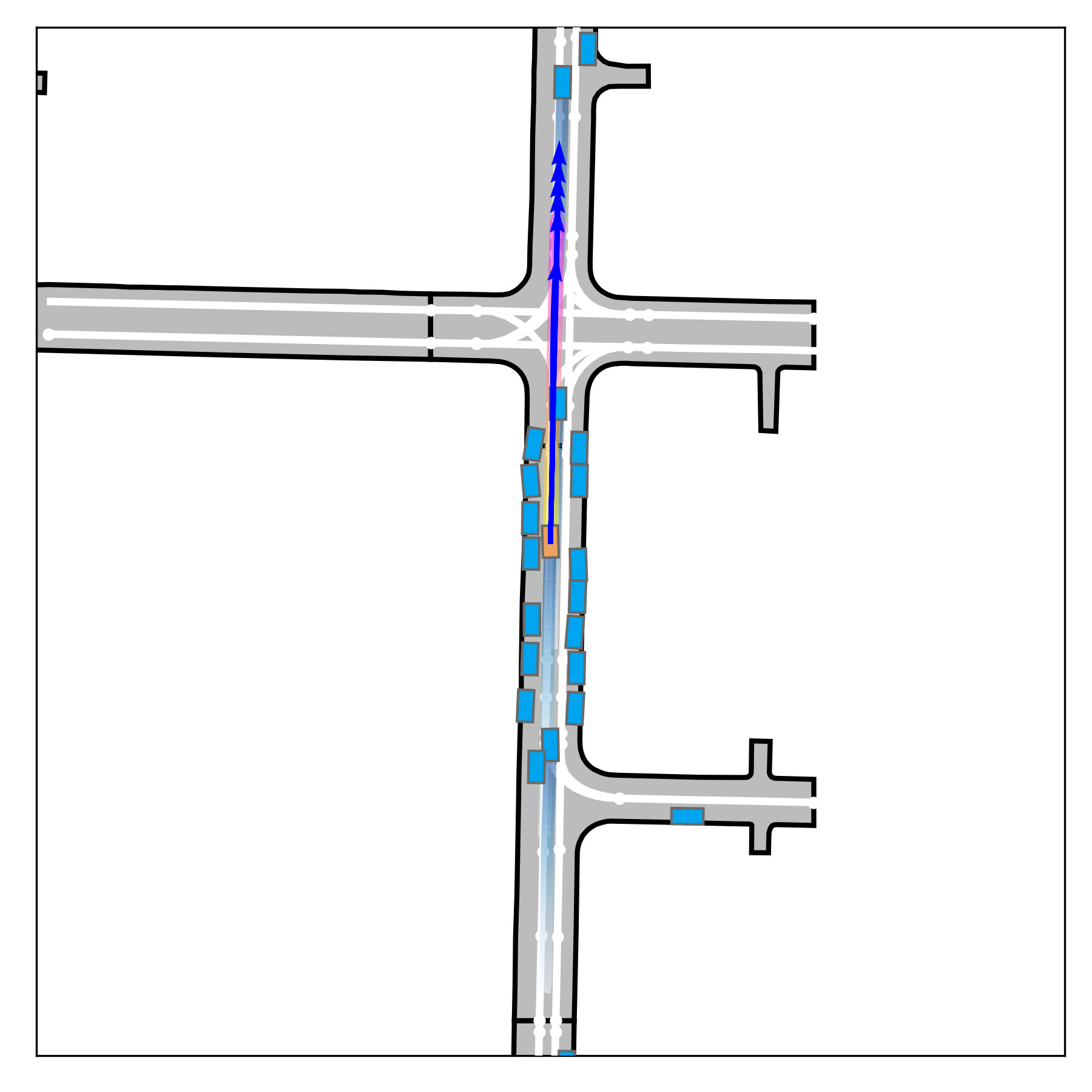} \hspace{-2.5mm}
    \includegraphics[width=0.24\textwidth]{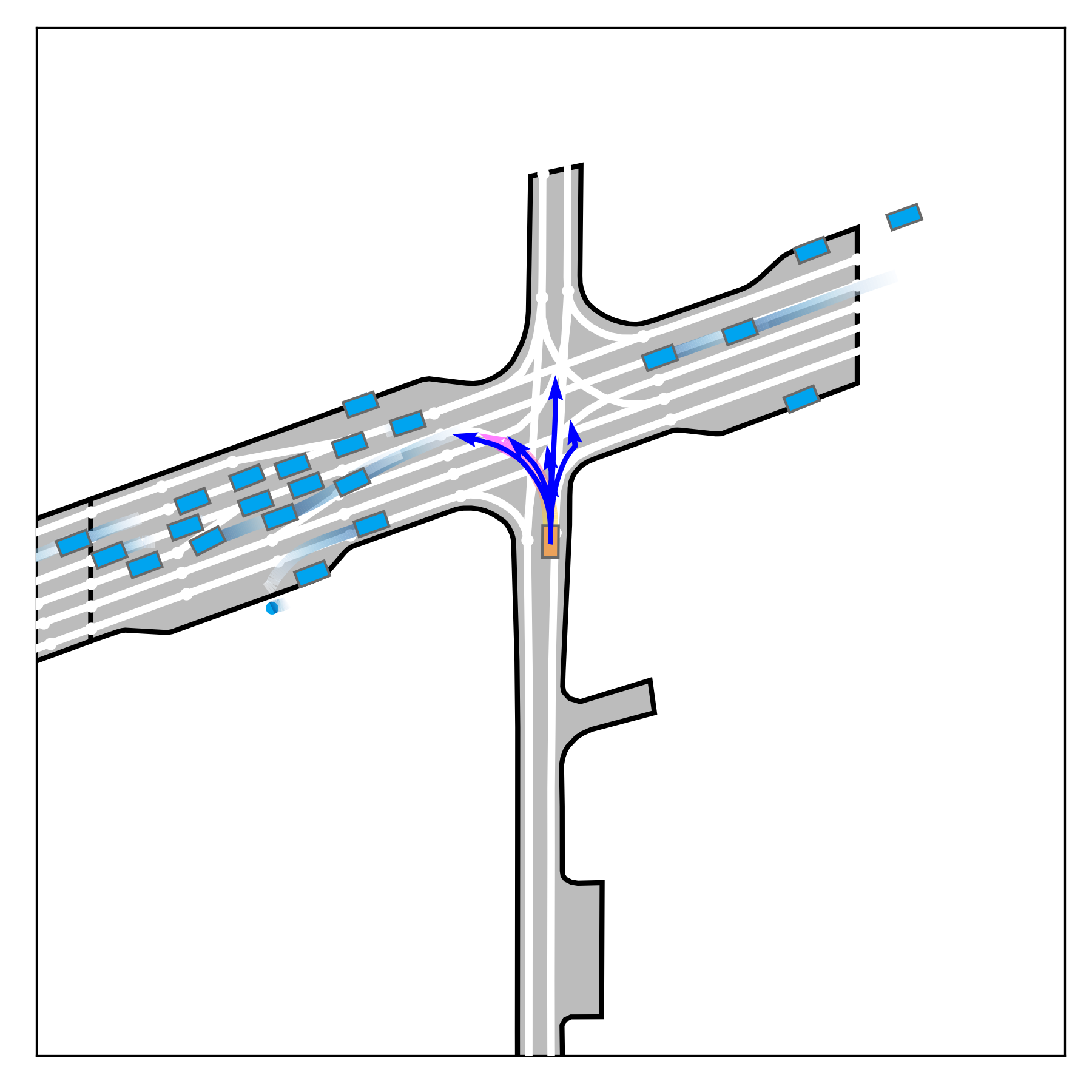} \hspace{-2.5mm}
    \includegraphics[width=0.24\textwidth]{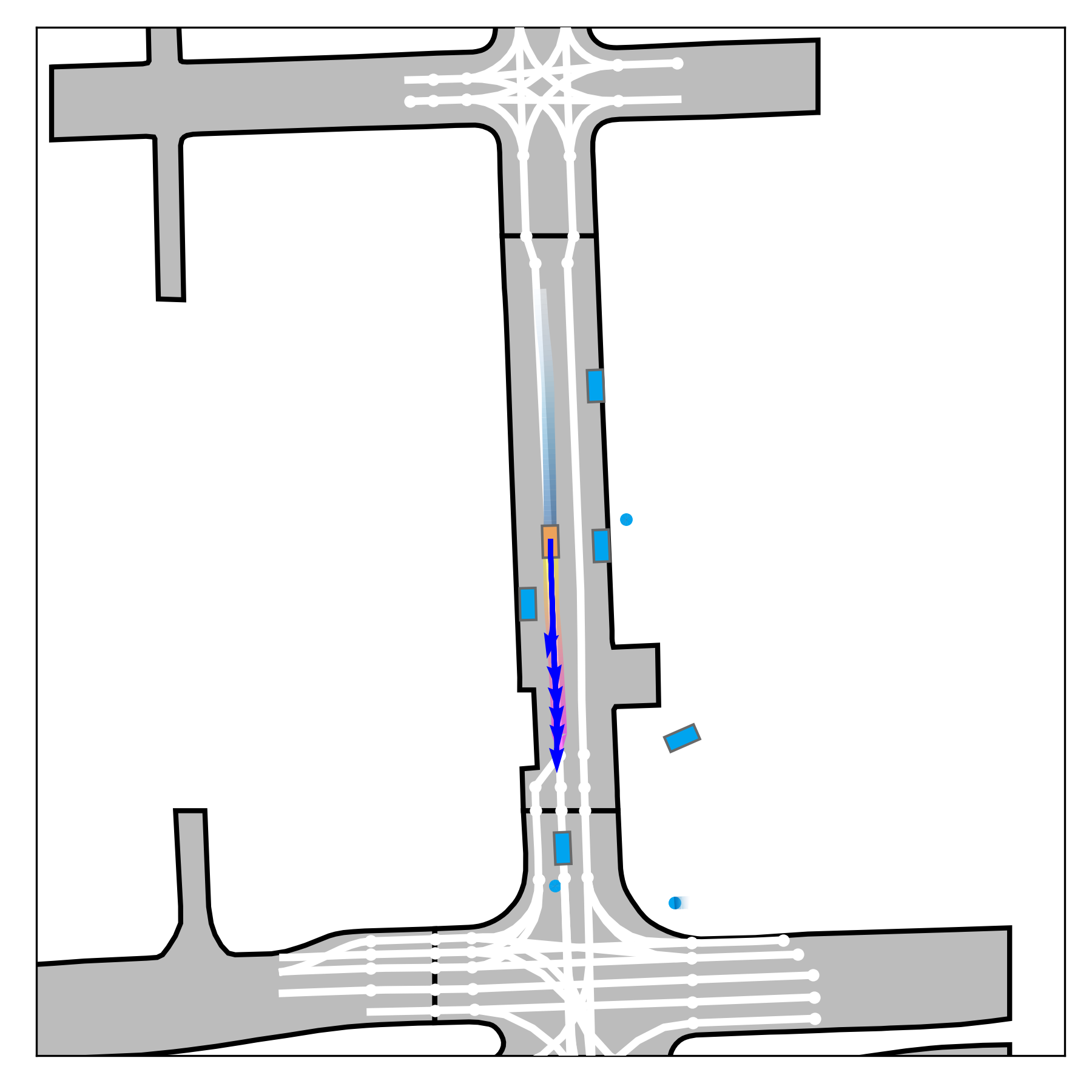} \hspace{-2.5mm}
    \includegraphics[width=0.24\textwidth]{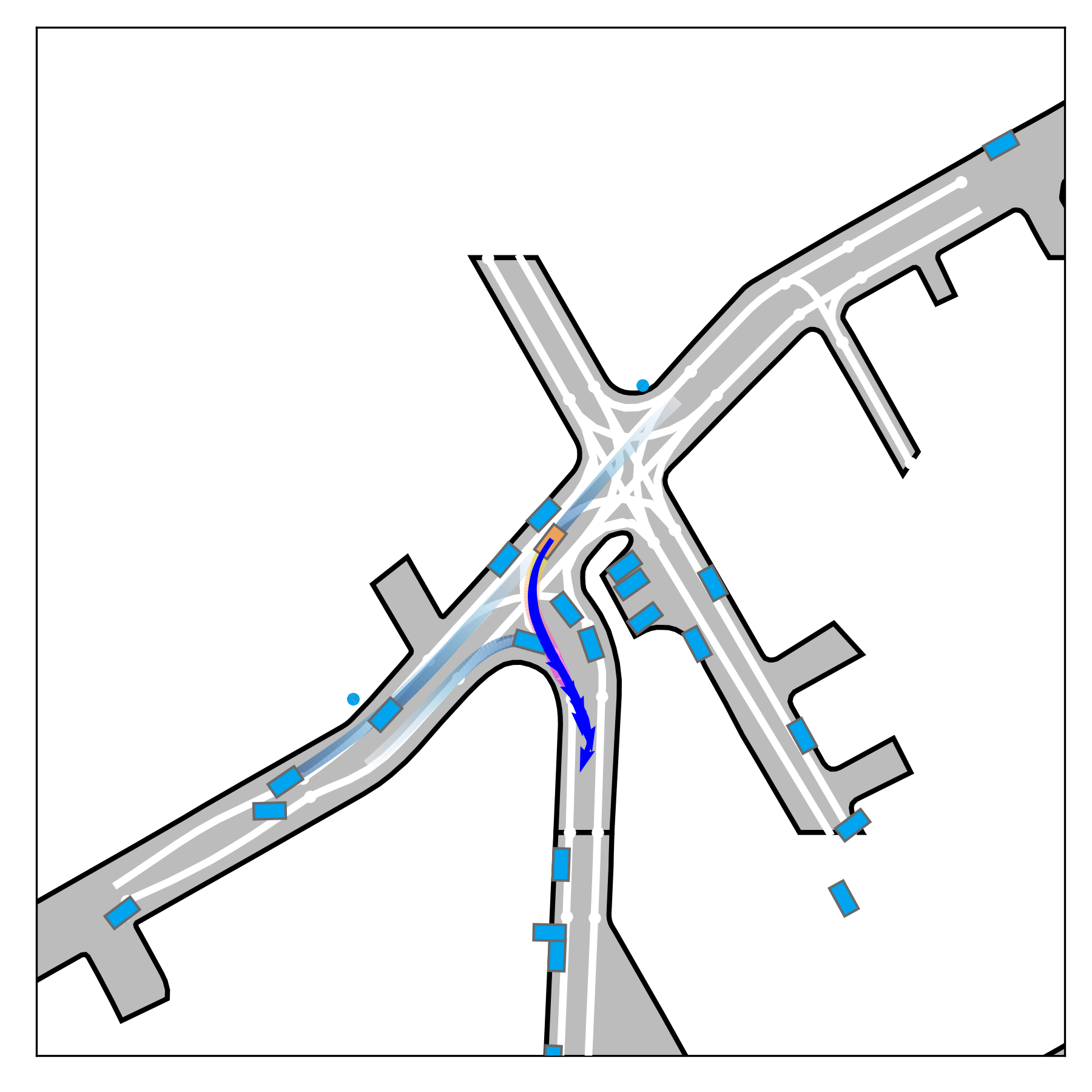} \hspace{-2.5mm}
    }
    \vspace{-3mm}
    \caption{ Qualitative results. We compare FINet to the state-of-the-art method, QCNet \citep{zhou2023query}. \textcolor{blue}{Blue arrows} represent the predicted future trajectories (K=6), while the \textcolor{pink}{pink arrow} denotes the ground truth future trajectory. The orange bounding box indicates the focal agent, while the blue bounding boxes denote surrounding agents. Compared to QCNet which usually cannot capture potential future trajectory pattern successfully, the proposed method demonstrates the ability to produce diverse (Columns 2) yet accurate (Columns 2,4) predictions. In certain scenarios, QCNet generates implausible predictions (Column 1, 3), which can be avoided by the proposed method. }
    \label{fig:vis}
    \vspace{-5mm}
\end{figure*}

\textbf{Effectiveness Analysis.} 
We compare the efficiency of the proposed method with the previous approach QCNet, which is a pure transformer-based method, as shown in Tab. \ref{tab:latency}. The results demonstrate that our method achieves overall better performance while significantly reduces model size by 52\% (3.7M vs. 7.7M), latency by 68\% (17.72ms vs. 54.55ms), GPU memory usage by 81\% (0.56G vs. 2.92G), and FLOPs by 95\% (1.47G vs. 28.0G), highlighting its superior efficiency. We attribute these improvements to the extensive use of the Mamba architecture, which scales linearly with sequence length. Notably, the reduction in FLOPs is significantly higher compared to other metrics. We hypothesize that this discrepancy arises because Mamba relies more on sequential computation, which is less optimized for highly parallelized hardware (e.g., GPUs), whereas attention mechanisms are fully optimized for parallel execution. As a result, real-world speed gains may not be directly proportional to theoretical FLOPs reductions. 
Some qualitative visualizations are presented in Fig. \ref{fig:vis} where the proposed method can produce more accurate and diverse future trajectories.


\subsection{Ablation Study}
In this section, we evaluate the impact of the decoder type and inductive bias. The experimental results are presented in Tab. \ref{tab:ablation_decoder}. Regarding the decoder type, we observe that the MLP-based decoder and the Query-based decoder achieve comparable performance when used with the proposed architecture. Specifically, the MLP-based decoder directly predicts future trajectories based on the output of FIM, whereas the Query-based decoder integrates learnable queries into TEDec rather than FIM.
In contrast, the proposed Intention-based decoder significantly outperforms both. We attribute this improvement to the integration of future trajectories into scene encoding. The joint optimization of all scene elements and future trajectories results in a more informative traffic representation. 

Additionally, we evaluate the impact of the inductive bias discussed in Sec. \ref{sec:fi_mamba}. Our findings indicate that omitting the inductive bias from future trajectory tokens or applying it to all trajectory tokens leads to suboptimal performance. The best results are obtained when the inductive bias is applied exclusively to the first future trajectory token. We hypothesize that, given the scanning mechanism of the Mamba block, this design facilitates the effective propagation of inductive bias, containing information related to the predicted endpoint, to subsequent future trajectory tokens, resulting in a more coherent and well-distributed set of future trajectories. In contrast, removing this mechanism would result in the absence of predicted endpoint information, while applying it to all tokens would excessively amplify its influence.
More ablation studies and implementation details can be found in the supplemental material.
\section{Conclusion}
In this work, we propose a highly efficient architecture named Future-Aware Interaction Network (FINet), an interaction-based approach for motion forecasting. FINet first models future trajectories in advance and then integrates them into scene encoding, ensuring mutual awareness among all elements. This approach enables the learning of a comprehensive traffic representation, leading to accurate and optimal trajectory predictions. To enhance both spatial and temporal modeling, we incorporate a State Space Model (SSM), specifically Mamba. To adapt Mamba for spatial interaction modeling, we introduce an adaptive reordering strategy that transforms unordered data into a structured sequence, enabling effective processing. Additionally, Mamba is also used to temporally refine future generated trajectories, ensuring consistent predictions.  We conducted extensive experiments on the widely used Argoverse 1 and Argoverse 2 datasets, demonstrating that the proposed method surpasses previous approaches in both performance and efficiency.

{
    \small
    \bibliographystyle{ieeenat_fullname}
    \bibliography{main}
}

\end{document}